\pdfoutput=1
\documentclass[11pt]{article}

\usepackage[preprint]{acl}
\usepackage{times}
\usepackage{latexsym}
\usepackage[T1]{fontenc}
\usepackage[utf8]{inputenc}
\usepackage{microtype}
\usepackage{inconsolata}
\usepackage{graphicx}
\usepackage{algpseudocode}
\usepackage{float}
\usepackage{tikz-dependency}

\title{LuxBank: The First Universal Dependency Treebank for Luxembourgish}

\author{Alistair Plum, Caroline Döhmer, Emilia Milano, \\ \textbf{Anne-Marie Lutgen, Christoph Purschke} \\
        University of Luxembourg \\ Esch-sur-Alzette, Luxembourg \\
        \texttt{\{alistair.plum,caroline.doehmer,emilia.milano\}@uni.lu} \\
        \texttt{\{anne-marie.lutgen,christoph.purschke\}@uni.lu}}
        
\begin{document}
\maketitle
\begin{abstract}
The Universal Dependencies (UD) project has significantly expanded linguistic coverage across 161 languages, yet Luxembourgish, a West Germanic language spoken by approximately 400,000 people, has remained absent until now. In this paper, we introduce LuxBank, the first UD Treebank for Luxembourgish, addressing the gap in syntactic annotation and analysis for this `low-research' language. We establish formal guidelines for Luxembourgish language annotation, providing the foundation for the first large-scale quantitative analysis of its syntax. LuxBank serves not only as a resource for linguists and language learners but also as a tool for developing spell checkers and grammar checkers, organising existing text archives and even training large language models. By incorporating Luxembourgish into the UD framework, we aim to enhance the understanding of syntactic variation within West Germanic languages and offer a model for documenting smaller, semi-standardised languages. This work positions Luxembourgish as a valuable resource in the broader linguistic and NLP communities, contributing to the study of languages with limited research and resources.
\end{abstract}

\section{Introduction}
The Universal Dependencies (UD) project has facilitated the production of treebanks across many languages, although some languages are still not represented almost 10 years after its original release \cite{nivre-etal-2016-universal}. With 161 languages represented as of the latest release, and a total of 283 treebanks across these languages, the  language coverage is undeniably vast.\footnote{Latest release at the time of writing: 15.05.2024.} The range of languages includes many of the major world languages, as well as varieties and dialects. However, some languages are still not represented at all, and Luxembourgish was one such case until recently. 

A West Germanic language closely related to German, Luxembourgish is spoken by roughly 400,000 people, mainly in Luxembourg \cite{Gilles2019}. Historically, Luxembourg has had a complex multilingual society where French and German have been predominantly used for official and formal (written) communication. In contrast, Luxembourgish was mostly a spoken language used informally between Luxembourgers until recently. With the rise of digital and social media, however, Luxembourgish has started to develop in the written domain and significant amounts of text data have started to become available, coupled with active language policies promoting Luxembourgish. Research in Natural Language Processing (NLP) for Luxembourgish has been limited until now, often in favour of French, German, and English. This has resulted in a situation where Luxembourgish is considered by some to be a `low-\textit{research}' language, as opposed to a low-\textit{resource} language.

In addition, large-scale syntactic annotation and analysis has not been undertaken before for Luxembourgish, making Luxembourg one of the few countries whose national language is not represented in the UD treebanks. This remains true despite the fact that four treebanks are available for Standard German \cite{volker2019hdt,mcdonald2013universal,zeman2018conll,basili2017toward}, as well as three non-standard treebanks for Swiss German \cite{aepli2018parsing}, Low Saxon \cite{siewert2021towards} and Bavarian \cite{Blaschke2024}. None of these represent a Middle-German variety, however, indicating an opportunity to extend the coverage for varieties of (or related to) German.

Aiming to address this gap in research, we present LuxBank, the first UD treebank for Luxembourgish. This project will be the first large-scale quantitative analysis of Luxembourgish syntax, and with this paper, we introduce the first formal guidelines for Luxembourgish language annotation. To this end, we present work related to Luxembourgish in Section \ref{sec:relwork} and describe the creation of LuxBank in Section \ref{sec:luxbank}, including highlighting notable syntactic phenomena. We discuss difficulties encountered in the creation process in Section \ref{sec:discussion} and conclude the paper with Section \ref{sec:conc}.

\section{Related Work}\label{sec:relwork}
Four UD treebanks exist for German, GSD \cite{mcdonald2013universal}, PUD \cite{zeman2018conll}, LIT \cite{basili2017toward} and the largest, HDT \cite{volker2019hdt}, at around 189k sentences. For non-standard varieties of German there are three UD treebanks: the UZH for Swiss German \cite{aepli2018parsing}, the LSDC for Low Saxon \cite{siewert2021towards} and as of recently, MaiBaam for Bavarian \cite{Blaschke2024}.

Two sets of guidelines for the UD project have been released since its inception, the first for version 1 \cite{nivre-etal-2016-universal} and the second for version 2 \cite{nivre2020universal}. As the current iteration of the project is version 2, we adhered to these guidelines, although we will discuss some aspects of the version 1 guidelines that could have been useful for our project in Section \ref{sec:discussion}.

\subsection{Luxembourgish Syntax}
Early work on the syntax of Luxembourgish can be found in \citet{schanen_recherche_1980} and in a few chapters of grammar books \cite{schanen_letzebuergesch_2012}. Certain characteristics of Luxembourgish syntax were later on investigated by dialectologists working on syntactic phenomena in West Germanic \cite{glaser_zur_2006} or presented in overview papers on Luxembourgish \cite{gilles_lux_2023}. A more in-depth analysis of syntactic features was conducted by \citet{doehmer2020}, and there are studies on neighbouring topics, namely pronominal reference for female persons \cite{martin_hatt_2019} and variation in inflectional morphology \cite{entringer_vun_2022}, but linguistics research on Luxembourgish syntax and on grammar in general is still in its beginnings. As there is relatively little research literature, we will invest more time into detecting, discussing, and categorising syntactic phenomena parallel to the annotation.

\subsection{Luxembourgish NLP}
Luxembourgish is underrepresented in NLP compared to its linguistic neighbours, French and German. Early research includes resources for NLP tasks \cite{adda-decker-etal-2008-developments}, analysis of writing patterns \cite{snoeren-etal-2010-study}, and a corpus for language identification \cite{lavergne-etal-2014-automatic}. Recent advancements feature sentiment analysis pipelines \cite{sirajzade-etal-2020-sentiment, Gierschek2022}, an orthographic correction pipeline \cite{purschke2020attitudes}, a zero-shot topic classification approach \cite{philippy-etal-2024-forget}, and automatic comment moderation \cite{ranasinghe-etal-2023-publish}. \textsc{LUX-ASR} provides Automatic Speech Recognition for Luxembourgish \cite{gilles-etal-2023-asrlux, gilles-etal-2023-luxasr}, while language models like \textsc{LuxGPT} leverage transfer learning from German \cite{bernardy2022}. Additionally, \textsc{LuxemBERT} matches multilingual BERT's performance in Luxembourgish tasks \cite{lothritz-etal-2022-luxembert, lothritz2023comparing}, and \textsc{ENRICH4ALL} supports a multilingual chatbot in administrative contexts \cite{anastasiou-2022-enrich4all}. While some tools and models exist for basic language processing, such as a limited spaCy integration\footnote{\url{https://github.com/PeterGilles/Luxembourgish-language-resources/blob/master/spaCy for Luxembourgish.ipynb}} and the python tool spellux for lemmatisation\footnote{\url{https://github.com/questoph/spellux}}, there is no published work on these tasks.

\begin{figure*}[ht!]
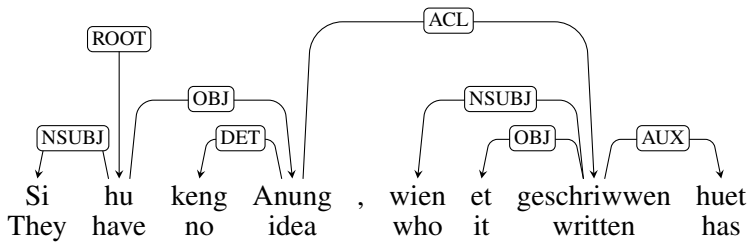

\centering
\begin{dependency}
    \begin{deptext}[column sep=0.4em]
    Si \& hu \& keng \& Anung \& , \& wien \& et \& geschriwwen \& huet \\
    They \& have \& no \& idea \& \& who \& it \& written \& has \\
    \end{deptext}
    \deproot{2}{ROOT}
    \depedge{2}{1}{NSUBJ}
    \depedge{4}{3}{DET}
    \depedge{2}{4}{OBJ}
    \depedge{8}{6}{NSUBJ}
    \depedge{8}{7}{OBJ}
    \depedge{4}{8}{ACL}
    \depedge{8}{9}{AUX}
\end{dependency}
\caption{Auxiliary verb in sentence c12.}
\label{fig:aux}
\end{figure*}

\section{LuxBank}\label{sec:luxbank}
In this section, we set out the methodology for the first round of annotations for LuxBank and reflect on specific linguistic conditions, such as standardisation and structural properties of Luxembourgish. The initial steps include translating the Cairo CICLing sentences, setting up preprocessing, as well as defining the annotation process. For the continuation of this project, we present the next steps in section \ref{sec:planned_work}, which are focused on adding further sentences from various domains of writing.

The project group working on LuxBank is made up of researchers from a range of different disciplines and specialisations: Two PhD researchers from the research project TRAVOLTA\footnote{\url{https://purschke.info/en/travolta}} with a background in linguistics, one expert for Luxembourgish grammar and syntax, and one computational linguist specialising in NLP for Luxembourgish. This is of central importance to our approach, as we are trying to incorporate computational processing and linguistic analysis on an equal footing in the development of the project. This is also due to the fact that linguistic experts are often underrepresented in computational linguistics projects. In the following, we describe the data annotation and analysis process. 

The Luxembourgish language is not fully standardised and presents a considerate amount of variation, be it lexical, grammatical, or phonological \cite{schnessen}. For this project, we decided to use written Luxembourgish according to the official spelling rules.\footnote{\href{https://portal.education.lu/Portals/79/Documents/WEB_LetzOrtho_Oplo5_v02-1.pdf?ver=2021-01-13-085421-963}{D’Lëtzebuerger Orthografie, 
Zenter fir d’Lëtzebuerger Sprooch (ZLS) 2022.}} Luxembourgish has an `emerging standard' and regional variants are being levelled. It is unclear whether there is significant syntactic variation stemming from the different dialects. Given the small size of the country and the ongoing efforts at standardisation, we argue that the variant of written Luxembourgish we are using comes very close to a standard language. The syntactic variation we find in the data is limited and can in most cases be explained through structural reasons.

For our first annotation set, we translate the 20 sentences from the Cairo CICLing corpus into Luxembourgish to ensure comparability. For the second round of annotations we will focus on news texts (journalistic language), as they represent a domain of formal writing and comply with the latest version of the spelling rules published in 2022.\footnote{\href{https://portal.education.lu/Portals/79/Documents/WEB_LetzOrtho_Oplo5_v02-1.pdf?ver=2021-01-13-085421-963}{D’Lëtzebuerger Orthografie, ZLS 2022.}} The choice of this specific written data is mainly due to practicality reasons, as those texts are easily accessible and offer a good starting point for the project. In the future, we will be open to add texts from different genres to cover a broader range of written language use in practice. 

\begin{table*}[ht!]
\centering
\begin{tabular}{|l|c|c|c|c|c|c|}
\hline
           & hunn & sinn & goen & ginn & kréien & wäert \\ 
           & (have) & (be) & (go) & (give) & (get) & (will) \\ \hline
main verb      & +     & +     & +     & +     & +     & –     \\ \hline
copula         & –     & +     & –     & +     & –     & –     \\ \hline
past tense     & +     & +     & –     & –     & –     & –     \\ \hline
passive voice  & –     & +     & –     & +     & +     & –     \\ \hline
subjunctive mood & –     & –     & +     & +     & –     & +/–   \\ \hline
future tense   & –     & –     & –     & –     & –     & +/–   \\ \hline
\end{tabular}
\caption{Functional properties of Luxembourgish auxiliary verbs, adapted from \citet{nubling_auf_2006} by \citet{doehmer2020}.}
\label{tab:verb_props}
\end{table*}

\subsection{CICLing Sentences}
The first 20 sentences are translated from the Cairo CICLing\footnote{\url{https://github.com/UniversalDependencies/cairo}} sentences, as recommended in the UD guide for submitting new treebanks.\footnote{\url{https://universaldependencies.org/release_checklist.html}} We use the English sentences as source language, and ask native speakers to perform the translations. We employ the available NLP resources for Luxembourgish to perform tokenisation, that is, the available Luxembourgish model for spaCy and spellux for obtaining lemmas.

Of note for our tokenisation is that we split contracted prepositions and determiners manually, which we adopt from Standard German. For the same reason we do not split hyphenated compound words. We deviate from the German guidelines with the determiner \textit{d'}, which does not exist in German, and for which we follow the French standard of tokenising it as \textit{d'}, therefore keeping the punctuation intact.

\subsection{Annotation} 
After the corpus selection, the two PhDs working on this project discuss each sentence. The discussion includes analysing the syntactic structure and dependencies by referring to the UD guidelines for German\footnote{\url{https://universaldependencies.org/de/}} and current work on Luxembourgish syntax \cite{doehmer2020}. The analysis starts by annotating the Part-of-Speech (POS) tags for every token. Then, the PhDs adhere to the classic UD process by starting with the main clause, detecting the root and its dependencies with the constituents of the clause. Afterwards, the secondary clause is the main focus of the discussion, looking at the connection with the main clause and its dependencies. Then, as a further step, the two linguists consult the syntactic expert for Luxembourgish to discuss their previous decisions, make additional changes and have a final validation of the dependency annotation.

The difficulties encountered during the annotation process mainly relate to the following reasons: First, the number of people available to work on this project is limited. Since Luxembourgish grammar is not taught in school, finding student assistants who could be trained as annotators is difficult; Second, the two PhDs working on the annotations have limited experience with UD annotation; and third, sometimes there is a missing overlap of Luxembourgish grammatical phenomena with the available UD tags. 

\subsection{Special Linguistic Features} \label{sec:special_linguistic_features}
In this section, we introduce the syntactic phenomena that need a more thorough explanation, as the tags offered by the UD are not sufficient to cover all the grammatical details unique to the Luxembourgish sentence structure. 

\subsubsection{The Verbal Domain}
We first focus on the verbal domain, describing the categorisation of different functional verb classes during the initial period of the project. 

\paragraph{Auxiliary Verbs}
As with most of the Germanic and Romance languages, Luxembourgish has a set of auxiliary verbs to serve different grammatical purposes, such as periphrastic constructions to express the past tense, subjunctive mood, or passive voice. In general, there are six auxiliaries in Luxembourgish, namely \textit{hunn, sinn, goen, ginn, kréien}, and \textit{wäert}, which can also occur as lexical verbs with the meaning of, respectively, `to have, to be, to go, to give, to get', with the exception of \textit{wäert} (`will') which has a defective paradigm and only works as a function verb. Each of these verbs, when used as an auxiliary, has a specific function, e.g. tense or mood. When used as main verb, these verbs are marked as \textit{root}, while, when used as auxiliaries, they are marked as \textit{aux}, together with modal verbs. Table \ref{tab:verb_props} summarises the functional properties of the Luxembourgish auxiliary system, and Figure \ref{fig:aux} shows an annotated sentence from LuxBank.

\begin{figure}[H]
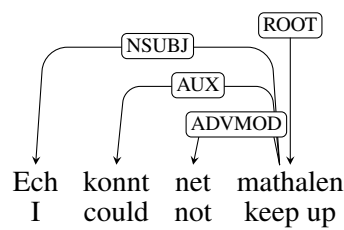

\centering
\begin{dependency}
   \begin{deptext}[column sep=0.4em]
     Ech \& konnt \& net \& mathalen \\
     I \& could \& not \& keep up \\
   \end{deptext}
   \deproot{4}{ROOT}
   \depedge{4}{2}{AUX}
   \depedge{4}{1}{NSUBJ}
    \depedge{4}{3}{ADVMOD}
\end{dependency}
\caption{Modal verb in sentence c18.}
\label{fig:mod}
\end{figure}

\paragraph{Modal Verbs}
Like other Germanic languages, Luxembourgish has a set of modal verbs that indicate the modality of the verbal phrase, i.e., if a situation/action is likely, possible, required etc. These are: \textit{kënnen, mussen, sollen, däerfen} and \textit{wëllen}, meaning, respectively, `can, must, shall, may, want'. Since there is no dedicated tag for modal verbs in the UD, this category too goes under the \textit{aux} tag. In some grammatical descriptions, they are referred to as `modal auxiliaries' \cite{Barbiers-van-Dooren}. Therefore, in LuxBank grammatical auxiliaries and modal verbs are marked with the same dependency tag. An annotated example from LuxBank is shown in Figure \ref{fig:mod}.

\begin{figure}[H]
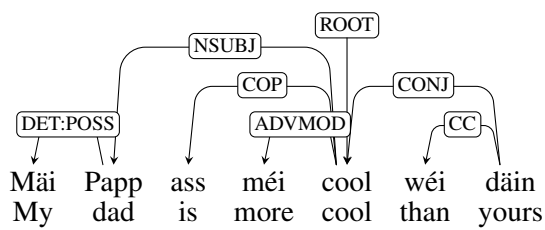

\centering
\begin{dependency}
   \begin{deptext}[column sep=0.5em]
     Mäi \& Papp \& ass \& méi \& cool \& wéi \& däin \\
     My \& dad \& is \& more \& cool \& than \& yours \\
   \end{deptext}
   \deproot{5}{ROOT}
   \depedge{5}{2}{NSUBJ}
   \depedge{5}{3}{COP}
   \depedge{2}{1}{DET:POSS}
   \depedge{5}{4}{ADVMOD}
   \depedge{7}{6}{CC}
   \depedge{7}{5}{CONJ}
\end{dependency}
\caption{Copular verb in sentence c8.}
\label{fig:cop}
\end{figure}

\paragraph{Copular Verbs}
It is worth underlining here that Luxembourgish, like many other Germanic languages, has more than one verb which can form a copular construction, e.g. \textit{ginn} (`to give') or \textit{sinn} (`to be'). As it is not possible to have more than one copular verb in the UD, at present, \textit{sinn} is registered as copula, while \textit{ginn} is only mentioned as an auxiliary. Figure \ref{fig:cop} shows an annotated example from LuxBank.

\begin{figure*}[ht!]
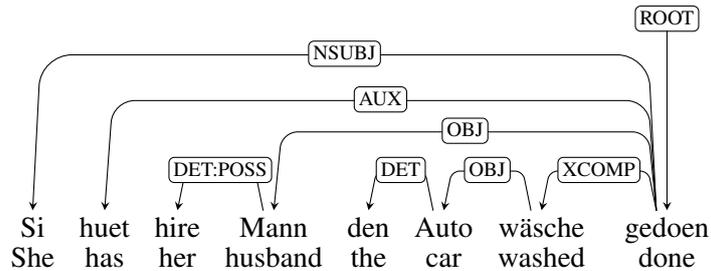

\centering
\begin{dependency}
   \begin{deptext}[column sep=0.4em]
     Si \& huet \& hire \& Mann \& den \& Auto \& wäsche \& [0.5em] gedoen \\
     She \& has \& her \& husband \& the \& car \& washed \& done \\
   \end{deptext}
   \deproot[edge unit distance=4ex]{8}{ROOT}
   \depedge[edge unit distance=1.7ex]{8}{1}{NSUBJ}
   \depedge[edge unit distance=1.4ex]{8}{2}{AUX}
   \depedge[edge unit distance=1.5ex]{8}{4}{OBJ}
   \depedge{4}{3}{DET:POSS}
   \depedge{8}{7}{XCOMP}
   \depedge{7}{6}{OBJ}
   \depedge{6}{5}{DET}
\end{dependency}
\caption{Causative verb in sentence c6.}
\label{fig:caus}
\end{figure*}

\begin{figure*}[ht!]
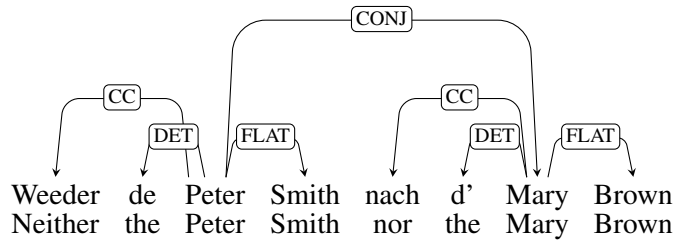

\centering
\begin{dependency}
   \begin{deptext}[column sep=0.4em]
     Weeder \& de \& Peter \& Smith \& nach \& d' \& Mary \& Brown \\
     Neither \& the \& Peter \& Smith \& nor  \& the \& Mary \& Brown \\
   \end{deptext}
   \depedge[edge start x offset=-6pt]{3}{1}{CC}
  \depedge{3}{2}{DET}
   \depedge{3}{4}{FLAT}
   \depedge{7}{5}{CC}
   \depedge{7}{6}{DET}
   \depedge{3}{7}{CONJ}
   \depedge{7}{8}{FLAT}
\end{dependency}
\caption{Determiner and proper name in sentence c11.}
\label{fig:prop_name}
\end{figure*}

\paragraph{Causative Verbs}
The verb \textit{doen} `to do' can be used to form a causative construction. Causatives indicate that a person or event is causing an action to happen. This auxiliary was already attested in Old and Middle High German \cite{Hans-Bianchi} and persists in Luxembourgish but not in Modern Standard German. However, the use of \textit{doen} is very selective towards its governed verbal phrase, as it can only be combined with specific main verbs. Its status is unclear because it has the functional and structural properties of an auxiliary but the semantic properties of a lexical verb. We tag it as \textit{root} to identify it as a lexical head rather than an auxiliary, considering its limited use and to maintain consistency within the under-specified auxiliary category. An annotated sentence featuring a causative verb is shown in Figure \ref{fig:caus}.

\subsubsection{The Nominal Domain}
When focusing on further syntactic elements, we find that Luxembourgish also shows a few structural peculiarities in the nominal domain which are worth mentioning.

\paragraph{Determiner and Proper Name}
A common phenomenon in Luxembourgish is the obligatory definite article before proper names. Like in any other noun phrase, the determiner is inflected based on number, gender, and case. Therefore, two or more dependencies in simple noun phrases are quite frequent, especially if the complete name of the person is mentioned. In these cases, we use the tag \textit{det} for the determiner, and following the UD guidelines, \textit{flat} for the second name or surname of the person. The annotated example sentence from LuxBank is shown in Figure \ref{fig:prop_name}.

\begin{figure}[H]
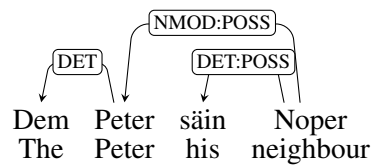
                               
\centering
\begin{dependency}
   \begin{deptext}[column sep=0.4em]
     Dem \& Peter \& säin \& Noper \\
     The \& Peter \& his \& neighbour\\
   \end{deptext}
   \depedge{4}{2}{NMOD:POSS}
   \depedge[edge start x offset=-6pt]{4}{3}{DET:POSS}
   \depedge{2}{1}{DET}
\end{dependency}
\caption{Possessive construction in sentence c7.}
\label{fig:posscon}
\end{figure}

\paragraph{Possessive Constructions}
The genitive is not an active case in the Luxembourgish language. Possessive relations can be expressed with an adnominal dative (only for animate possessors) or with a \textit{vun}-PP \cite{doehmer2020}. An annotated example sentence is shown in Figure \ref{fig:posscon}.

\begin{figure*}[ht!]
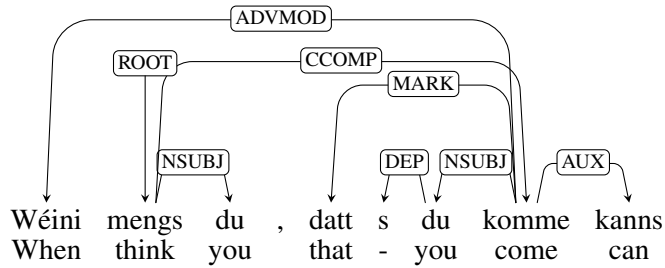

\centering
\begin{dependency}
    \begin{deptext}[column sep=0.4em]
    Wéini \& mengs \& du \& , \& datt \& s \& du \& komme \& kanns \\
    When \& think \& you \&      \& that \& - \& you  \& come \& can \\
    \end{deptext}
    \depedge{2}{3}{NSUBJ}
    \depedge[edge unit distance=1.8ex]{2}{8}{CCOMP}
    \depedge[edge unit distance=2ex]{8}{1}{ADVMOD}
    \depedge{8}{5}{MARK}
    \depedge{8}{7}{NSUBJ}
    \depedge{8}{9}{AUX}
    \depedge{7}{6}{DEP}
    \deproot{2}{ROOT}
\end{dependency}
\caption{Agreement marker in sentence c14.}
\label{fig:clit}
\end{figure*}

\subsubsection{Other Domains}
Since not every phenomenon in Luxembourgish can be analysed with the UD tagset, we decided to use the miscellaneous attributes for the annotation to explicate the phenomena. The miscellaneous attributes, labelled in the MISC column, are intended for the annotators to put in additional information about a tag.\footnote{\url{https://universaldependencies.org/misc.html}} At the moment, there are two phenomena that are covered by this tag, the negation and the agreement marker, described in the tag set as \textit{s} clitic. 

\paragraph{Negation}
The negation in Luxembourgish is typically expressed as a negation particle with \textit{net}. In the first version of the UD tagset, the negation was a proper tag, but in the second version the tag is no longer available and is now tagged as \textit{advmod}. We will use the feature \textit{Polarity=NEG} for the negation particle, as is the custom in other UD treebanks.

\paragraph{Agreement Marker}
In subordinate clauses, where the subject is the second person singular (\textit{du/de}), the complementiser is followed by the agreement marker \textit{s}. The \textit{s}-marker is mandatory in this sentence structure and has an orthographically isolated position between the initial element of the subordinate clause and the \textit{du/de}-pronoun \cite{doehmer2020}. It developed out of a reanalysis of the inflectional (verbal) \textit{s}-suffix (2nd person singular) and became a clitic before the subject pronoun. Over time it grammaticalised into an obligatory \textit{s}-marker with a fixed syntactic position. As there is no available tag to properly describe this phenomenon, we decided to use the \textit{dep} tag and describe it in the miscellaneous column with \textit{clitic}. In general, this is not a case of clitic doubling as in some West Germanic dialects because the subject pronoun itself is not always used as a clitic. Moreover, the \textit{s}-clitic appears after any element in the complementiser position, not only subordinating conjunctions, but also after interrogative phrases or long prepositional phrases \cite{doehmer2020}. Therefore, it should not be linked to the complementiser. Given the fact that it is syntactically bound and very predictable in terms of the sentence type in combination with a specific subject pronoun, attaching it to the verb with the \textit{expl} relation (as per the UD guidelines) would not be justified. Although it doesn't behave like a regular clitic, the \textit{clitic}-tag seems to be the most suitable, because of the strong dependence on the subject pronoun \textit{du/de}. This phenomenon has different structural properties in the Continental West Germanic varieties (it doesn't appear in other standard languages, though) and the terminology may vary in some descriptions \cite{Renkwitz}. 

Figure \ref{fig:clit} gives an example sentence from LuxBank where this phenomenon is annotated.

\subsection{Planned Work}\label{sec:planned_work}
Extending the coverage of LuxBank is our primary objective, with the next batch of sentences currently being annotated. This batch comprises 50 randomly sampled sentences\footnote{Sentences longer than 25 tokens were not considered.} from news articles from RTL, the main news broadcaster of Luxembourg. For further extensions, we plan to translate sentences from xSID \cite{vandergoot2021maskedlanguagemodelingtranslation} to support comparability across further NLP tasks in various languages. While working on this extension, we will also add the morphological features in the initial and future set of sentences.

\section{Discussion}\label{sec:discussion}
After applying the UD guidelines and analysing the Luxembourgish sentences, we now discuss practical and theoretical aspects related to the syntactic structure of the 20 CICLing sentences, including under-specified tags and potential challenges when incorporating different languages. Although the CICLing sentences are drawn from simple everyday language, the analysis of such sentences can be quite complex, e.g., when they contain elliptic constructions. Ellipses are a common phenomenon in many European languages, but it is difficult to determine syntactic dependencies, when different parts of the sentence have been elided. Among the 20 CICLing sentences, at least five contain some sort of elliptical structure. As a consequence, CICLing corpus might not be the best starting point for developing new treebanks, since some of the fundamental basic syntactical structures are not as well represented.  

\begin{figure*}[ht]
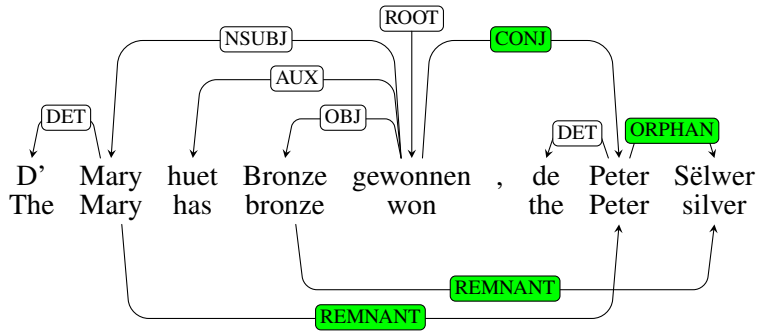

\centering
\begin{dependency}
    \begin{deptext}[column sep=0.4em]
    D' \& Mary \& huet \& Bronze \& gewonnen \& , \& de \& Peter \& Sëlwer \\
    The \& Mary \& has \&  bronze \& won \&  \& the  \& Peter \& silver \\
    \end{deptext}
    \depedge{5}{2}{NSUBJ}
    \depedge[edge unit distance=1.8ex]{8}{7}{DET}
    \depedge[edge unit distance=2ex,label style={fill=green!}]{8}{9}{ORPHAN}
    \depedge{2}{1}{DET}
    \depedge{5}{3}{AUX}
    \depedge{5}{4}{OBJ}
    \depedge[label style={fill=green!}]{5}{8}{CONJ}
    \deproot{5}{ROOT}
    \depedge[edge below, edge unit distance=1ex, label style={fill=green!}]{4}{9}{REMNANT}
    \depedge[edge below, edge unit distance=1.2ex, label style={fill=green!}]{2}{8}{REMNANT}
\end{dependency}
\caption{UD v2 versus v1 (below) annotation of ellipsis in sentence c9.}
\label{fig:ell_v2}
\end{figure*}

To better understand their structure, we analyse the sentences with elliptical structure following both the UD guidelines of version 1 and version 2, see the respective syntactical analysis in Figure \ref{fig:ell_v2}. Although the version 2 UD guidelines are currently in use, where the dependency between the head of the elliptic sentence and the element depending on the omitted verb is marked as \textit{orphan}, we find the version 1 to be more accurate from a linguistic point of view. In a verb phrase ellipsis, connecting the two \textit{nsubj} under the tag \textit{remnant} and leaving the other dependencies unvaried (i.e. as the verb phrase were there) would better reflect the underlying structure of these sentences. 

A further discrepancy between linguistic theories and UD guidelines, as already mentioned in \ref{sec:special_linguistic_features}, concerns the \textit{aux} tag. This tag is under-specified and used for two classes of functional verbs: auxiliaries and modal verbs. While the miscellaneous column can be helpful to deal with the limits of the UD guidelines in practice, it is still a makeshift solution that does not do full justice to phenomena not yet covered by the guidelines. As the feature column is still not enough to distinguish between different verb classes, a dedicated tag to allow better differentiation between auxiliary and modal verbs would be more precise from a linguistic point a view. Moreover, limiting the classification to a single copular verb further reduces the linguistic accuracy of the UD. The possibility to add more than one copular verb would then result in a more realistic representation of the class of copular verbs in Luxembourgish, without compromising the comparability with other languages.

Another aspect regarding the CICLing sentences concerns the modeling of gendered languages. As English usually does not mark the grammatical gender of common nouns, languages with marked gender then need to decide on the grammatical gender of these nouns. Although this is not strictly related to the syntactic dependencies in the sentence, it could lead to a different interpretation and therefore an inaccurate translation of the original sentence. The following example from the CICLing sentences (c7) illustrates this:

\begin{itemize}
    \item[(EN)] Peter's \textbf{neighbour} painted the fence red.
    \item[(DE)] \textbf{Der Nachbar} von Peter hat den Zaun rot (an)gemalt.
    \item [(LB)] Dem Peter \textbf{säin Noper} huet den Zonk rout ugestrach.
\end{itemize}

As can be seen in the example sentences (marked in bold), even if the grammatical gender is unmarked in English, in both target languages the translators chose the male version of the word, arguably perpetuating the unaware gender bias of male and female roles in society \cite{bolukbasietal}. While we do not foresee cases like this in future additions to LuxBank, since we will be using original Luxembourgish material instead of translations, we feel it is important to point this out.

LuxBank is an ongoing project and the main goal is to add more annotated sentences to the treebank. Since this is the beginning of the project, we are continuously adapting the guidelines for Luxembourgish while annotating the data. More linguistic features for Luxembourgish will need to be specified in the future, as they weren't covered in the initial 20 sentences, e.g., loanwords, verb cluster variation, and doubly filled complementisers.

Given the amount of language contact phenomena in Luxembourgish, especially loanwords from German, French, or English are a frequently occurring phenomenon that needs to be addressed. In the nominal domain, further guidelines must be created for French and English compounds, aside from using the \textit{flat} tag, as they are sometimes written as one word, as separate units, or hyphenated, depending on either the spelling norms of the source language or on Luxembourgish orthography.\footnote{\href{https://portal.education.lu/Portals/79/Documents/WEB_LetzOrtho_Oplo5_v02-1.pdf?ver=2021-01-13-085421-963}{D’Lëtzebuerger Orthografie, ZLS 2022.}} French compounds often appear as multi-word units and are therefore close to syntactic expressions \cite{VanGoethemAmiot2019}. Some of those expressions are directly borrowed into Luxembourgish, e.g. \textit{Projet de loi} `bill (draft law)' or \textit{Carte d'identité} `identity card'. These expressions will need to be tagged according to French morphology and left-headedness. It should also be avoided that the French prepositions \textit{de} and \textit{d'} are automatically tagged as Luxembourgish definite articles.

Another common pattern in Luxembourgish syntax is verb cluster variation. The order of elements in 2-, 3-, and 4-verb clusters is variable, when modal verbs or subjunctive auxiliaries appear in subordinate clauses \cite{doehmer2020}. In general, word order variation will not affect the deep structure of the sentence, i.e., the dependencies remain the same, but the surface structure will be different. Concerning the left periphery of subordinate clauses, the initial element of the subordinate clause is sometimes extended by a second complementiser, namely \textit{dass/datt} \cite{doehmer2020}. Sentences with a doubly filled complementiser, such as \textit{obwuel dass et reent} `(lit.) although that it rains', could cause difficulties in the annotation process because in most cases the complementiser position can only contain a single constituent. All of these phenomena (among others) have to be addressed in the future to develop appropriate guidelines for Luxembourgish.

\section{Conclusion}\label{sec:conc}
In this paper, we introduce LuxBank as the first treebank for Luxembourgish. As the discussion of structural characteristics and challenges encountered when developing annotation guidelines for Luxembourgish show, building a new treebank for a small language represents a theoretical as well as practical challenge. This is particularly true in view of the structural variation in Luxembourgish and its ongoing standardisation. In this context, the decision to bring together a mixed team of linguistic and computational experts has proven crucial to the successful implementation of UD for Luxembourgish.

LuxBank will facilitate a more in-depth understanding of Luxembourgish as a `low-research' language, making it an invaluable resource not only for linguists but also for language teaching. This treebank project can serve as an aid for spell-checking tools as well as for future grammar checking applications. A tailor-made tagging system derived from earlier versions of LuxBank could ensure higher accuracy and consistency in Luxembourgish text processing and modelling, to help to better organise existing text archives, and to extend the treebank further. In the future, LuxBank will enable easier quantitative exploration of linguistic data, providing insights that were previously more difficult to obtain.

From a typological perspective, it is important to complete the data in the UD treebanks for West Germanic varieties. So far, mainly large standard languages have been incorporated, whereas regional varieties and/or smaller languages are underrepresented. LuxBank adds the first Middle German language description to the UD. This can help to explore syntactic variation and to understand the structural aspects of these languages. 

LuxBank will also be beneficial for NLP research and text processing in general. Presently, the support for Luxembourgish is limited to certain tasks (lemmatisation, POS), and the available resources do not use the UD tagset for POS tagging. Building a dedicated treebank for Luxembourgish will make it possible to extend the support for the language in industry-standard tools like \textit{spaCy} to the grammatical level and to offer a comparable tag set for the analysis of syntactic structures. In doing so, LuxBank is laying the foundation for a better representation of Luxembourgish in NLP, both for further research and for the development of customized tools and pipelines.

Luxembourgish can also serve as a model case for describing other small languages and varieties, as these often possess unique characteristics – and resulting challenges – like those discussed in this paper: a limited amount of available resources, a small number of trained linguistic experts, a high amount of linguistic variation (be it lexical, grammatical, or orthographic), a structural influence from other (standard) languages, and a complex multilingual language situation. With this contribution, we aim to position Luxembourgish as a valuable resource for comparable language situations. We also hope to highlight the importance of foundational research for small and non-standardised languages to preserve linguistic diversity in the digital age and make it more visible in NLP.

\section*{Limitations}
The work presented in this paper is still in progress, and subsequent modifications may be made as the project evolves. It is important to note that finding and recruiting domain experts for data annotation is challenging. Additionally, the amount of variation within the language sometimes makes it difficult to reach a consensus on the classification of phenomena, which has introduced additional complexity to our research.

\section*{Ethics Statement}
All data used in this project is freely available and obtained from publicly accessible sources. The human annotators involved in this project were fully compensated for their contributions, as this work forms part of their regular employment responsibilities. Additionally, all data is appropriately licensed for the intended use in this research, ensuring compliance with legal and ethical standards. This adherence to ethical guidelines ensures the integrity and responsible conduct of our research.

\section*{Acknowledgements}
This research was supported by the Luxembourg National Research Fund (Project code: C22/SC/117225699).

\bibliography{custom}

\end{document}